\documentclass[letterpaper, 10 pt, conference]{bib/ieeeconf}  

\IEEEoverridecommandlockouts                              

\usepackage{slstyle}

\title{\LARGE \bf
Towards Search-based Motion Planning for Micro Aerial Vehicles}
\author{Sikang Liu$^{1}$,  Kartik Mohta$^{1}$, Nikolay Atanasov$^{2}$, and Vijay Kumar$^{1}$
\thanks{This work is supported by ARL \# W911NF-08-2-0004, DARPA \# HR001151626/HR0011516850, ARO \# W911NF-13-1-0350, and ONR \# N00014-07-1-0829.}
\thanks{$^{1}$S. Liu, K. Mohta, and V. Kumar are with the GRASP Laboratory, University of Pennsylvania, USA
 {\tt\small sikang@seas.upenn.edu}}%
\thanks{$^{2}$N. Atanasov is with the Electrical and Computer Engineering department, UC San Diego, USA
 {\tt\small natanasov@ucsd.edu}}%
}

\begin{document}

\maketitle
\thispagestyle{empty}
\pagestyle{empty}

\begin{abstract}
Search-based motion planning has been used for mobile robots in many applications. However, it has not been fully developed and applied for planning full state trajectories of Micro Aerial Vehicles (MAVs) due to their complicated dynamics and the requirement of real-time computation. In this paper, we explore a search-based motion planning framework that plans \emph{dynamically feasible}, \emph{collision-free}, and \emph{resolution optimal and complete} trajectories. This paper extends the search-based planning approach to address three important scenarios for MAVs navigation: (i) planning safe trajectories in the presence of motion uncertainty; (ii) planning with constraints on \emph{field-of-view} and (iii) planning in dynamic environments. We show that these problems can be solved effectively and efficiently using the proposed search-based planning with motion primitives.
\end{abstract}

\section{Introduction}
\label{sec:introduction}
Micro Aerial Vehicles (MAVs) are small multi-rotor helicopters that are able to freely fly in constrained and complex environments. It has been shown in~\cite{mellingerICRA2011} that the MAV dynamics are \textit{differential flat} which implies that the control inputs can be computed as functions of the flat outputs and their derivatives. Many works~\cite{mellingerICRA2011, hehn2011quadrocopter, mueller2015, richter2016polynomial} show precise control of MAVs through trajectories represented as time-parameterized piecewise polynomials with respect to these flat outputs. However, generating those trajectories is a challenging task considering that they have to be \emph{dynamically feasible}, \emph{collision-free}, and \emph{optimal}. Moreover, navigation of the MAV in unknown environments requires fast re-planning for avoiding new obstacles. Some recent works~\cite{oleynikova2016continuous, gao2017iros, liu2017planning} try to apply optimization algorithms that satisfy these requirements, however, these approaches require heuristic based initial guess to set up the optimization problem and are not complete. In addition, in more complex situations where motion uncertainty, limited \emph{field-of-view} (FOV) and moving obstacles exist, it is much harder to find the optimal result using the optimization-based approaches.

As shown in our previous work~\cite{liu_iros_2017}, we formulate the MAV planning problem as an optimization problem which is solvable through search-based methods using motion primitives. The curse of dimensionality is one of the major problems of search-based methods that prevent their use for high order dynamical systems. However, through the \emph{differential flatness} and discretization in control space (instead of state space), we are able to achieve fast computation in an induced lattice space and solve for trajectories that are \emph{dynamically feasible}, \emph{collision-free}, \emph{resolution complete}, and \emph{optimal}. \cite{liu_planning_2018}~shows a variation of this approach by modifying the constraint function for planning in SE(3). In this paper, we explore its potential to solve more practical planning problems. We mainly consider the following three challenges in real-world navigation tasks:
\begin{enumerate}
\item \emph{Planning with Motion Uncertainty}: The robot is not able to perfectly track the nominal trajectory in the presence of disturbances. We consider how to plan a safer trajectory that is less likely to crash the robot in an obstacle-cluttered environment.
\item \emph{Planning with Limited FOV}: Vision-based state estimation and the limited FOV of sensors to detect obstacles require the robot to travel with constraints on the yaw angle. We propose a way to find the desired yaw profile along the trajectory that obeys this constraint.
\item \emph{Planning in Dynamic Environments}: It is PSPACE-hard to plan trajectories taking into account moving obstacles~\cite{reif1994motion}. We consider a tractable approach to solving this problem with polynomial time complexity and further adopt it to solve the multi-robot planning problem.
\end{enumerate}

Our contribution in this work is to provide solutions to these problems using the search-based planning paradigm. In the following sections, we will discuss each of the above problems and demonstrate our solutions with results. The corresponding code can be found in the planning library at \url{https://github.com/sikang/mpl_ros}.

\section{General Problem Formulation}
\label{sec:problem_formulation}
Before introducing specific planning problems, we describe the general form of planning an optimal trajectory using search-based method. Similar to~\cite{mellingerICRA2011}, we select flat outputs of the MAV system as $\sigma = \begin{bmatrix}\mathrm{x} & \mathrm{y} & \mathrm{z} & \psi\end{bmatrix}^\T$. Let $x(t) \in \mathcal{X} \subset \mathbb{R}^{q m}$ be the state of the MAV system which includes its position in $\mathbb{R}^m$ and its $(q-1)$ derivatives (velocity, acceleration, \etc). Denote $\mathcal{X}^{free}\subset\mathcal{X}$ as the free region of the state space that consists of two parts: collision-free positions $\mathcal{P}^{free}$ and dynamical constraints $\mathcal{D}^{free}$, \ie maxim 	um velocity $\bar{v}_{max}$, acceleration $\bar{a}_{max}$, and higher order derivatives along each axis of $\sigma$. Therefore, 
\begin{equation}
\begin{aligned}
\mathcal{X}^{free} &= \mathcal{P}^{free} \times \mathcal{D}^{free},\\
\mathcal{D}^{free} &= [-\bar{v}_{max},\, \bar{v}_{max}]^m \times [-\bar{a}_{max},\, \bar{a}_{max}]^m \times \cdots.
\end{aligned}
\end{equation}

The obstacle regions and dynamically infeasible space is denoted as $\mathcal{X}^{obs} =\mathcal{X} \setminus \mathcal{X}^{free}$. 

The expression for the dynamical system is obtained as
\begin{gather}\label{eq:sys}
\dot{x}(t) = A x(t) + B u(t), \\
A = \begin{bmatrix}
    \mathbf{0} & \mathbf{I}_m  & \cdots &\mathbf{0}\\
    \vdots& \vdots &\ddots&\vdots\\
    \mathbf{0}& \cdots  & \mathbf{0} & \mathbf{I}_m\\
    \mathbf{0}& \cdots  & \mathbf{0} & \mathbf{0}\\
    \end{bmatrix}_{qm\times qm}, \quad B = \begin{bmatrix} \mathbf{0}\\\vdots\\\mathbf{0}\\\mathbf{I}_m\end{bmatrix}_{qm\times m},\nonumber
\end{gather}
and the control input is $u(t) \in \mathcal{U} = [-u_{max},u_{max}]^m \subset \mathbb{R}^m$. 

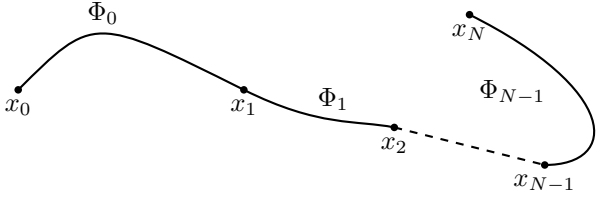
\begin{figure}[tb]
\centering
\begin{tikzpicture}[auto, node distance=2cm,>=latex']
\tikzset{joint/.style={circle, inner sep=0pt, minimum size=0.1cm, fill=black, draw=none}}
   \draw[thick] (0,1) node[joint] {} node[below] {$x_0$} .. controls (1,2) .. node[above] {$\Phi_0$}(3,1)
                (3,1) node[joint] {} node[below] {$x_1$}.. controls (4, 0.5) and (4.5, 0.6) .. node[above] {$\Phi_1$}(5, 0.5)  node[below] {$x_2$};
                
   \draw[thick, dashed] (5, 0.5) node[joint] {} -- (7, 0);
   \draw[thick] (7,0) node[joint] {}  node[below] {$x_{N-1}$}.. controls (8,0) and (8,1) .. node[near end] {$\Phi_{N-1}$} (6,2) node[below] {$x_N$} node[joint] {};
   
\end{tikzpicture}
\caption{A piecewise polynomial trajectory with $N$ segments.}
\label{fig:traj}
\end{figure}

A piecewise polynomial trajectory $\Phi$ with $N$ segments as shown in \autoref{fig:traj} is defined as:
\begin{equation}\label{eq:traj}
\Phi(t) =\left\{\begin{array}{lll}
                  \Phi_0(t-t_0),\ \ &t_0 \leq t < t_1,\\
                  \Phi_1(t-t_1),\ \ &t_1 \leq t < t_2,\\ 
                  \vdots\\
                  \Phi_{N-1}(t-t_{N-1}),\ \ &t_{N-1} \leq t \leq t_N.
                \end{array}
              \right.
\end{equation}
Each segment $\Phi_n$ is derived as the polynomial generated by applying the control input $u(t)$ on the state $x_n$ for a duration $\Delta t_n = t_{n+1} - t_{n}$ as per~\eqref{eq:sys}.

We define the \emph{smoothness} or \emph{effort} of a trajectory \wrt its $q$-th derivative as the square $L^2$-norm of the control input:
\begin{equation}
\label{eq:J}
J_q(\Phi) =  \int\limits_0^{T} \|\Phi^{(q)}\|^2 dt =  \sum_{n = 0}^{N-1}\int\limits_0^{\Delta t_n} \left\|u_n(t)\right\|^2 dt,
\end{equation}
where $u_n$ correspond to the control input for $n$-th segment.
An optimal trajectory (not guaranteed to be unique) that respects the dynamical and collision constraints, and is \emph{minimum-time} and \emph{smooth} can be obtained from:

\begin{problem}
\label{prob:1}
Given an initial state $x_0 \in \mathcal{X}^{free}$ and a goal region $\mathcal{X}^{goal} \subset \mathcal{X}^{free}$, find a polynomial trajectory $\Phi(t)$ such that:
\begin{equation}
  \label{eq:problem1}
  \begin{gathered}
    \argmin_{\Phi} \; J_q(\Phi) + \rho_T T\\
    \begin{aligned}
      \text{s.t.}\ \ &\dot{x}(t) = Ax(t)+Bu(t), \\
      &x(0) = x_0, \quad x(T) \in \mathcal{X}^{goal},\\
      &x(t) \in \mathcal{X}^{free},\quad u(t)\in{\mathcal{U}}.
    \end{aligned}
  \end{gathered}
\end{equation}
where  the parameter $\rho_T \geq 0$ determines the relative importance of the trajectory duration $T$ versus its smoothness $J_q$.
\end{problem}

As shown in~\cite{liu_iros_2017},~\prob\ref{prob:1} can be converted into a search problem using motion primitives that are generated from a finite set of constant control inputs $\mathcal{U}_M$. In the following sections, we use a graph search algorithm such as A* to compute an optimal solution for~\prob\ref{prob:1}. Details on motion primitive construction and graph search are omitted in this article due to space limitations but can be found in~\cite{liu_iros_2017}.

\section{Planning with Motion Uncertainty}
\label{sec:disturbance}
Existing work in trajectory planning assumes the availability of high control authority allowing robots to perfectly track the generated trajectories. However, this assumption is impractical in the real world since unpredictable environmental factors such as wind, air drag, wall effects can easily disturb the robot from the nominal trajectory. Thus, even though a nominal trajectory is in free space, it can easily lead to a collision when the robot gets close to obstacles. To reduce this risk, a trajectory that stays away from obstacles is desired. Traditionally, this is worked around by inflating the obstacle by a radius that is much larger than the actual robot size. However, this over-inflation strategy is not a complete solution for motion planning in obstacle-cluttered environments since it is prone to block small gaps such as doors, windows, and narrow corridors.

The reachable set (funnel)~\cite{tedrake2010lqr} is used to model motion uncertainty for a robot following a time-varying trajectory. Assuming bounded and time-invariant disturbances leads to bounded funnels. It is straightforward to show that the funnel of a linear system, as in~\eqref{eq:sys}, controlled by a PD-controller~\cite{lee2010geometric} is bounded by a certain radius with respect to the control gains. However, the planning strategy in~\cite{tedrake2010lqr} treats the motion uncertainty as a hard constraint for collision checking which is an over conservative strategy that discards all the trajectories close to obstacles. Besides, it is computationally expensive to search using funnels.

Alternatively, Artificial Potential Fields (APFs) are used to plan paths that are away from obstacles efficiently~\cite{warren1989, adeli2011path, li2013effective}. APFs have been used to model collision costs in trajectory generation through line integrals~\cite{ratliff2009chomp, oleynikova2016continuous, gao2017iros, usenko2017real} in which the safe trajectory is refined from an initial nominal trajectory through gradient descent. However, this gradient-based approach strongly relies on the initial guess of time allocation and the sampling of end derivatives for fast convergence and it ignores the dynamical constraints during the re-optimization. Moreover, the result is easily trapped in undesired local minima. Thus, it is not an appropriate method to solve the safe planning problem in complex environments.

In this section, we propose a novel approach that models the motion uncertainty as a soft constraint and plans for trajectories that are as safe as possible with respect to the collision cost through the line integral of the APF. The resulting trajectory is constrained to be within a tunnel from the initial trajectory, such that it is suitable for planning in unknown environments. The proposed approach does not require the Jacobian and Hessian of the cost functions and hence is computationally efficient.

\subsection{Problem Formulation}
We call the trajectory derived from solving~\prob\ref{prob:1} that ignores the collision cost as the nominal trajectory $\Phi_0$.
We treat trajectory planning with motion uncertainty as a problem of finding a locally optimal trajectory around the nominal $\Phi_0$ that takes into account the collision cost. It can be formulated as a variation of~\prob\ref{prob:1} where we add a collision cost $J_c$ in the objective function and a search region (tunnel) $\mathcal{T}(\Phi_0)$ around $\Phi_0$ in the constraints:
\begin{problem}
\label{prob:2}
Given an initial state $x_0 \in \mathcal{X}^{free}$, a goal region $\mathcal{X}^{goal} \subset \mathcal{X}^{free}$ and a search region $\mathcal{T}(\Phi_0)$ around the nominal trajectory $\Phi_0$, find a polynomial trajectory $\Phi$ such that:
\begin{equation}
  \label{eq:problem2}
  \begin{gathered}
    \argmin_{\Phi} \; J_q + \rho_T T + \rho_c J_c\\
    \begin{aligned}
      \text{s.t.}\ \ &\dot{x}(t) = Ax(t)+Bu(t)\\
      &x(0) = x_0, \; x(T) \in \mathcal{X}^{goal}\\
      &x(t) \in \mathcal{T}(\Phi_0) \cap \mathcal{X}^{free},\quad u(t)\in{\mathcal{U}}
    \end{aligned}
  \end{gathered}
\end{equation}
where the weights $\rho_T,~\rho_c \geq 0$ determines the relative importance of the trajectory duration $T$ and collision cost $J_c$ versus its smoothness $J_q$.
\end{problem}

In this section, we show that~\prob\ref{prob:2} can be converted into a search problem and solved using motion primitives.


\subsubsection{Collision Cost $J_c$}
We define the collision cost in~\prob\ref{prob:2} as the line integral:
\begin{equation}\label{eq:Jc}
J_c(\Phi) = \int_{\Phi} U(s)\,ds.
\end{equation}
where $U(s)$ is the potential value of position $s\in \mathbb{R}^m$ that is defined as:
\begin{equation}\label{eq:potential}
U(s) =\left\{\begin{array}{ll}
                  0,\ \ &d(s) \geq d_{\text{thr}}\\
                  F(d(s)),\ \ &d_{\text{thr}} > d(s) \geq 0
                \end{array}\right.
\end{equation}
where $d(s)$ is the distance of position $s$ from the the closest obstacle.
In addition, for positions that are away from obstacles more than a distance $d_{\text{thr}}$, we consider their collision cost to be negligible. Thus, the potential function $U(s)$ should be a non-negative and monotonically decreasing function in domain $[0, d_{\text{thr}})$ and equal to zero when $d \geq d_{\text{thr}}$. One choice for $F(\cdot)$ is an polynomial function with order $k > 0$:
\begin{equation}\label{eq:F}
F(d) = F_{\text{max}}\left(1 - \frac{d}{d_{\text{thr}}}\right)^k.
\end{equation}

\begin{figure}[tb]
  \centering
  \begin{subfigure}[b]{0.48\linewidth}
    \begin{tikzpicture}
      \node[above right] (img) at (0,0) {\includegraphics[width=\linewidth]{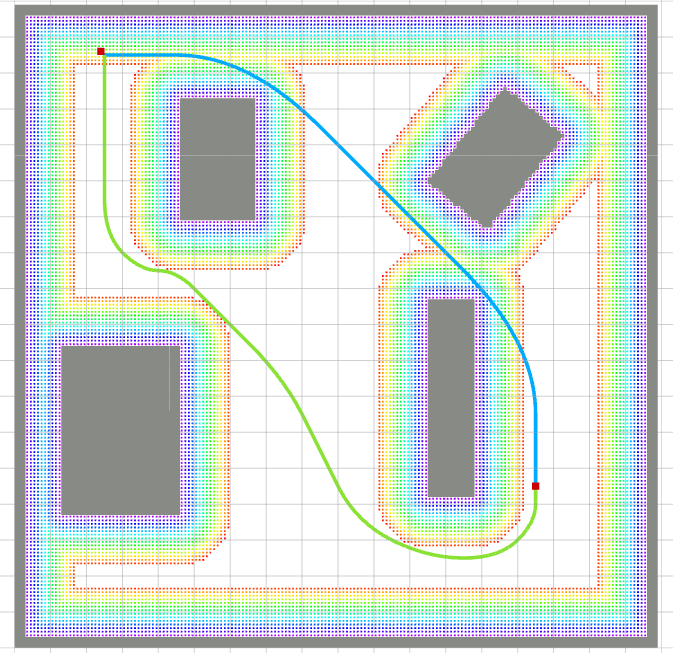}};
      \node at (100pt,25pt) {start};
      \node at (20pt,100pt) {goal};
    \end{tikzpicture}
    \caption{Global plans.}
  \end{subfigure}
  \begin{subfigure}[b]{0.48\linewidth}
    \begin{tikzpicture}
      \node[above right] (img) at (0,0) {\includegraphics[width=\linewidth]{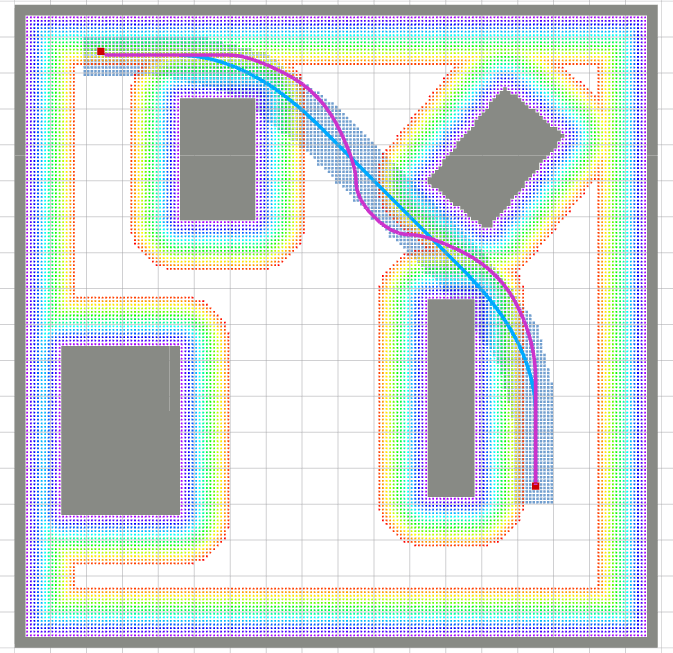}};
      \node at (100pt,25pt) {start};
      \node at (20pt,100pt) {goal};
    \end{tikzpicture}
    \caption{Local plans.}
  \end{subfigure}
  \caption{Planning in an occupancy grid map. Rainbow dots indicate the truncated \emph{Artificial Potential Field} (APF) generated from~\eqref{eq:potential}. In the left figure, the blue trajectory is the shortest trajectory that ignores collision cost; the green trajectory is the shortest trajectory that treats the APF as obstacles. In the right figure, the magenta trajectory is the planned trajectory using the proposed method that takes into account the collision cost. It is locally optimal within the tunnel (blue region) around the nominal shortest trajectory from (a).  \label{fig:traj_perturb}}
\end{figure}
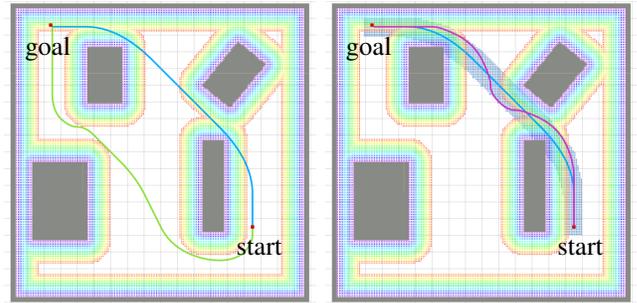

The analytic expression of the line integral in~\eqref{eq:Jc} is hard to compute, instead we sample the trajectory at $I$ points with uniform time step $dt$ for approximation:
\begin{equation}\label{eq:Jc2}
\int_{\Phi}U(s)ds \approx \sum_{i = 0}^{I-1} U(p_i)\|v_i\|dt
\end{equation}
where $dt = \frac{T}{I-1}$ and $p_i,v_i$ are corresponding position and velocity at time $i\cdot dt$. This approximation can be easily calculated when the obstacle and potential field are represented as a grid as shown in~\fig\ref{fig:traj_perturb}.


\subsubsection{Tunnel Constraint $\mathcal{T}$}

A tunnel is a configuration space around the nominal trajectory $\Phi_0$ that is used to bound the perturbation. Let $D(r)$ be the disk with radius $r$, the tunnel  $\mathcal{T}(\Phi_0, r)$ is the Minkowski sum of $D$ and $\Phi$ as:
\begin{equation}
  \mathcal{T}(\Phi_0, r) = \Phi_0 \oplus D(r).
\end{equation}

Note that $\mathcal{T}$ could overlap with obstacles. Thus, we enforce the valid state to be inside the intersection of $\mathcal{T}$ and free space $\mathcal{X}^{free}$ to guarantee safety. When $r\rightarrow \infty$,~\prob\ref{prob:2} is equivalent to computing a globally optimal trajectory.


\subsection{Solution}\label{sec:sampling}
Given the set of motion primitives $\mathcal{U}_M$ and the induced space discretization, we can reformulate~\prob\ref{prob:2} as a graph-search problem similar to~\cite{liu_iros_2017} which is solvable through dynamic programming algorithms such as Dijkstra and A*. For each primitive $\Phi_n$, we sample $I_n$ points to calculate its collision cost according to~\eqref{eq:Jc2}. In the grid map, the $I_n$ should be dense enough to cover all the cells that $\Phi_n$ traverses. One choice of automatically selecting $I_n$ is
\begin{equation}
I_n = \frac{\bar{v}_{max}\cdot \Delta t_n}{r_M}, ~\bar{v}_{max} = \max\{v_x, v_y, v_z\}.
\end{equation}
where $r_M$ is the grid resolution.

\subsection{Experimental Results}\label{sec:exp_perturb}
\begin{figure}[tp]
  \centering
   \begin{subfigure}[b]{0.49\linewidth}
        \includegraphics[width=\linewidth]{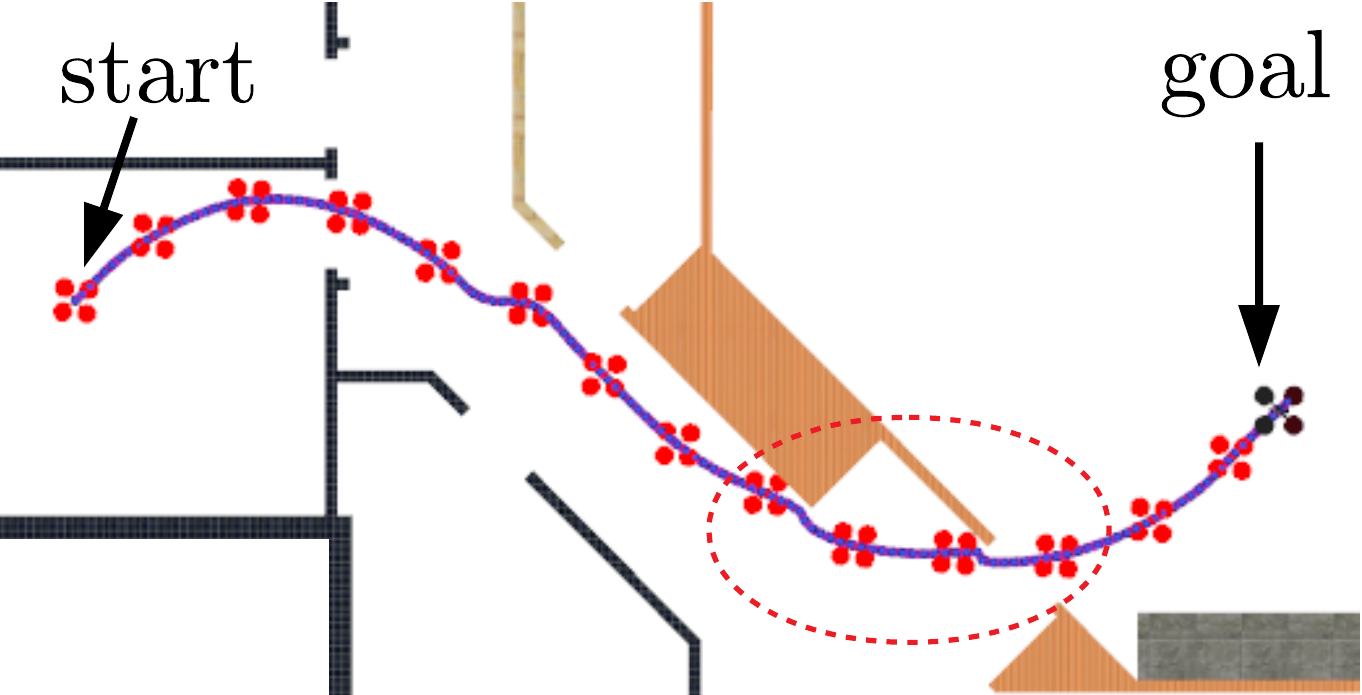}
        \caption{Without collision cost.}\label{fig:exp1}
    \end{subfigure}
       \begin{subfigure}[b]{0.49\linewidth}
        \includegraphics[width=\linewidth]{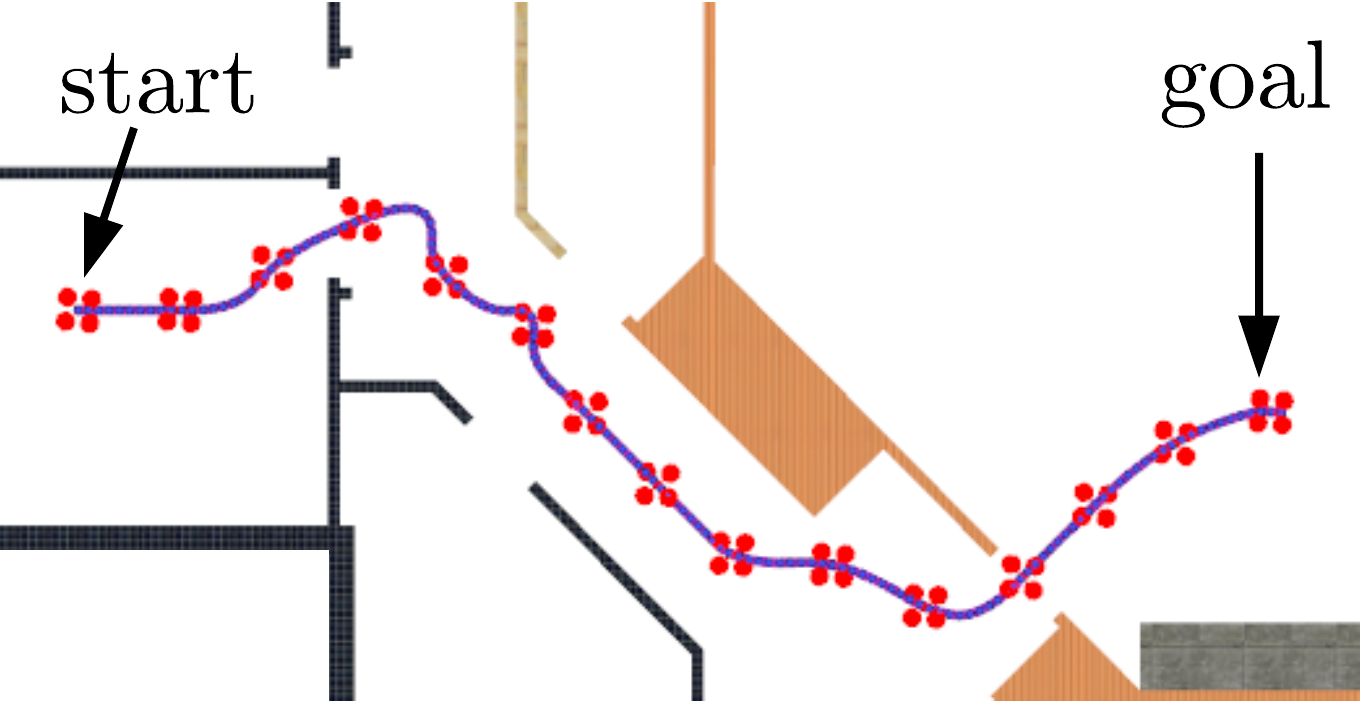}
         \caption{With collision cost.}\label{fig:exp2}
    \end{subfigure}
   \caption{MAV with limited sensing navigating in an office environment. The left and right figures show the results from using two different planners: (a) method in~\cite{liu_iros_2017} that does not consider collision cost; (b) the proposed method that plans for optimal trajectories with respect to collision cost. In (a), the robot touches the wall multiple times in the circled region. The trajectory in (b) is much safer. \label{fig:traj_perturb_exp}}
\end{figure}

In~\fig\ref{fig:traj_perturb_exp}, a quadrotor tries to reach the goal position using the proposed planner in an office environment. The environment is shown as a 2D colored schematic, but the robot initially has no information about the environment. Therefore, it needs to constantly re-plan at certain frequency to avoid new obstacles that appear in the updated map.~\fig\ref{fig:exp1} shows the results using traditional method in~\cite{liu_iros_2017} that doesn't consider collision costs, in which the quadrotor occasionally touches the wall inside the circled region.~\fig\ref{fig:exp2} shows the results from using the proposed method, in which the robot stays away from walls and safely goes in and out of rooms through the middle of open doors. The re-planning time using our method is fast in this 2D scenario, the run time of which is below \SI{10}{ms} for a \nth{2} order dynamic model.

\section{Planning with Limited FOV}
\label{sec:yaw}
Due to the fact that the yaw of a MAV system does not affect the system dynamics, this flat output is frequently ignored in existing planning works. Except when using omni-directional sensors, a fully autonomous MAV system is usually directional. In order to guarantee safety while navigating in an unknown environment, the MAV should always move in the direction that can be seen by a range sensor such as RGB-D or \emph{time-of-flight} (TOF) camera which has limited FOV. Thus, the yaw $\psi(t)$ should change as the robot moves. Specifically, the desired $\psi$ is related to the velocity direction: $\xi = \arctan{v_y/v_x}$. This constraint is non-linear and couples the flat outputs $\mathrm{x}$, $\mathrm{y}$, and $\psi$, thus it is hard to model it in the optimization framework as proposed in~\cite{richter2016polynomial} and~\cite{deits2015computing}. In this section, we develop a search-based method that resolves this constraint properly by splitting it into two parts: a soft constraint that minimizes the difference between $\psi$ and $\xi$ and a hard constraint that enforces the moving direction $\xi$ to be inside the FOV of the range sensor.

\subsection{Problem Formulation}
We define an additional cost term representing a soft FOV constraint as the integral of the square of angular difference between velocity direction and desired yaw:
\begin{equation}
J_{\psi}(\Phi) = \int\limits_{0}^{T}[\psi(t) - \xi(t)]^2 dt,
\end{equation}
while the hard constraint can be formulated by the absolute angular difference and the sensor's horizontal FOV $\theta$:
\begin{equation}
|\psi(t) - \xi(t) | \leq \frac{\theta}{2},
\end{equation}

To add these constraints, we modify~\prob\ref{prob:1} as follows:
\begin{problem}
\label{prob:3}
Given an initial state $x_0 \in \mathcal{X}^{free}$, a goal region $\mathcal{X}^{goal} \subset \mathcal{X}^{free}$ and a sensor FOV $\theta$, find a polynomial trajectory $\Phi$ such that:
\begin{equation}
  \label{eq:problem3}
  \begin{gathered}
    \argmin_{\Phi} \; J_q + \rho_T T + \rho_\psi J_{\psi}\\
    \begin{aligned}
      \text{s.t.}\ \ &\dot{x}(t) = Ax(t)+Bu(t)\\
      &x(0) = x_0, \; x(T) \in \mathcal{X}^{goal}\\
      &x(t) \in \mathcal{X}^{free},\quad u(t)\in{\mathcal{U}}\\
      &|\psi(t) - \xi(t)| \leq \frac{\theta}{2}
    \end{aligned}
  \end{gathered}
\end{equation}
where the weights $\rho_T,~\rho_\psi \geq 0$ determine the relative importance of the trajectory duration $T$, the yaw cost $J_\psi$, and its smoothness $J_q$.
\end{problem}

\subsection{Solution}
Since both of the additional constraints contain $\xi$ which is an $\arctan$ function, it is difficult to get their analytic expressions. We use a sampling method similar to the one in \eqref{eq:Jc2} to approximate the FOV constraint. The control $u_\psi \in u$ for yaw can be applied in a different control space compared to the other flat outputs. To be specific, we set $u_\psi$ as the angular velocity assuming the robot does not need to aggressively change the heading.

\begin{figure}[tb]
  \centering
   \begin{subfigure}[b]{0.23\textwidth}
   \begin{tikzpicture}
\node[above right] (img) at (0,0) {\includegraphics[width=1.00\textwidth]{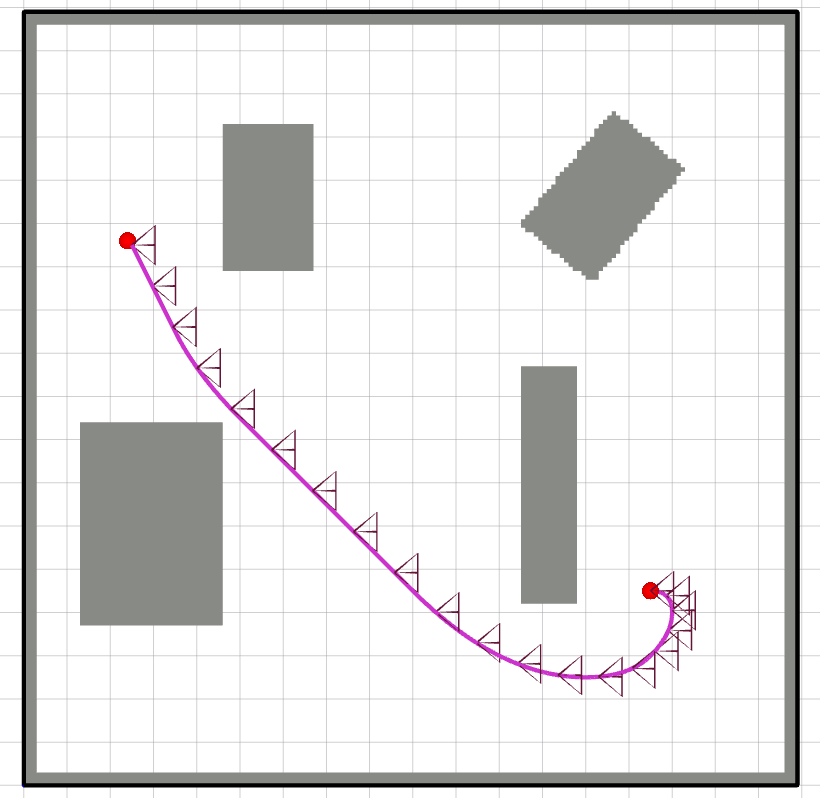}};
\node at (100pt,25pt) {start};
\node at (20pt,95pt) {goal};
\draw[->](100pt,33pt)--(113pt,33pt);
\end{tikzpicture}
   \caption{$\rho_\psi = 0,~ \theta = 2\pi$.}\label{fig:yaw0}
\end{subfigure}
    \begin{subfigure}[b]{0.23\textwidth}
       \begin{tikzpicture}
\node[above right] (img) at (0,0) {\includegraphics[width=1.00\textwidth]{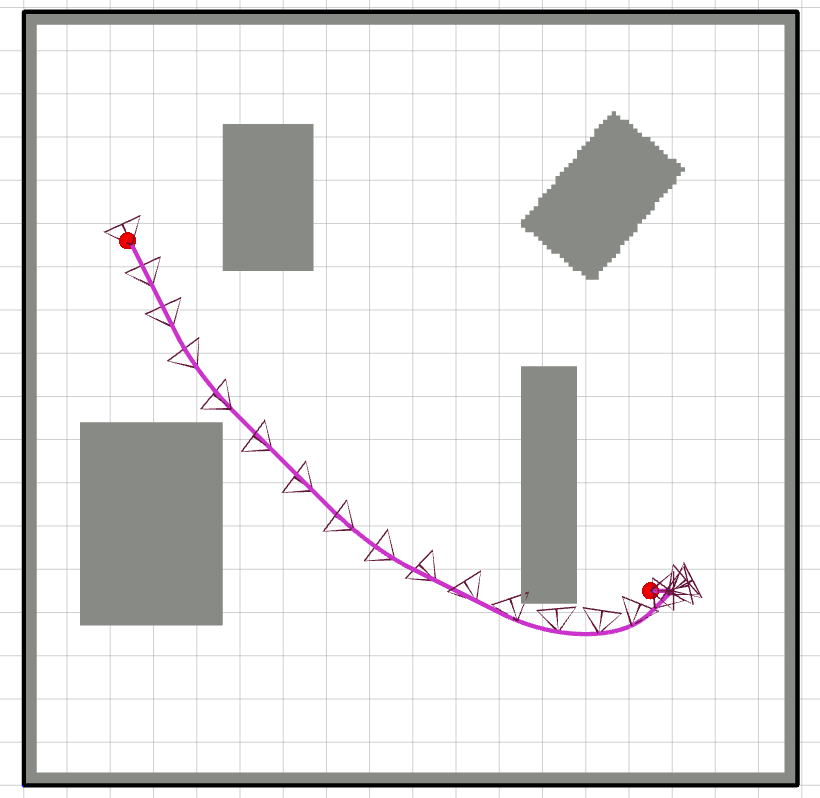}};
\node at (100pt,25pt) {start};
\node at (20pt,95pt) {goal};
\draw[->](100pt,33pt)--(113pt,33pt);
\end{tikzpicture}
     \caption{$\rho_\psi = 1,~ \theta = 2\pi$.}\label{fig:yaw1}
    \end{subfigure}
    \begin{subfigure}[b]{0.23\textwidth}
       \begin{tikzpicture}
\node[above right] (img) at (0,0) {\includegraphics[width=1.00\textwidth]{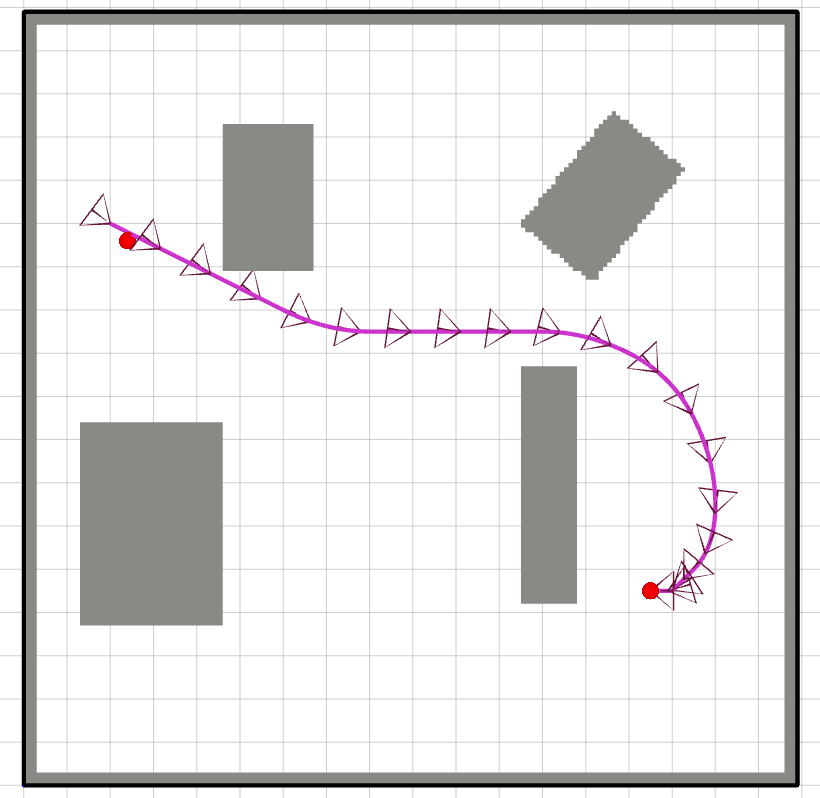}};
\node at (100pt,25pt) {start};
\node at (20pt,95pt) {goal};
\draw[->](100pt,33pt)--(113pt,33pt);
\end{tikzpicture}
        \caption{$\rho_\psi = 0,~ \theta = \pi/2$.}\label{fig:yaw2}
    \end{subfigure}
    \begin{subfigure}[b]{0.23\textwidth}
       \begin{tikzpicture}
\node[above right] (img) at (0,0) {\includegraphics[width=1.00\textwidth]{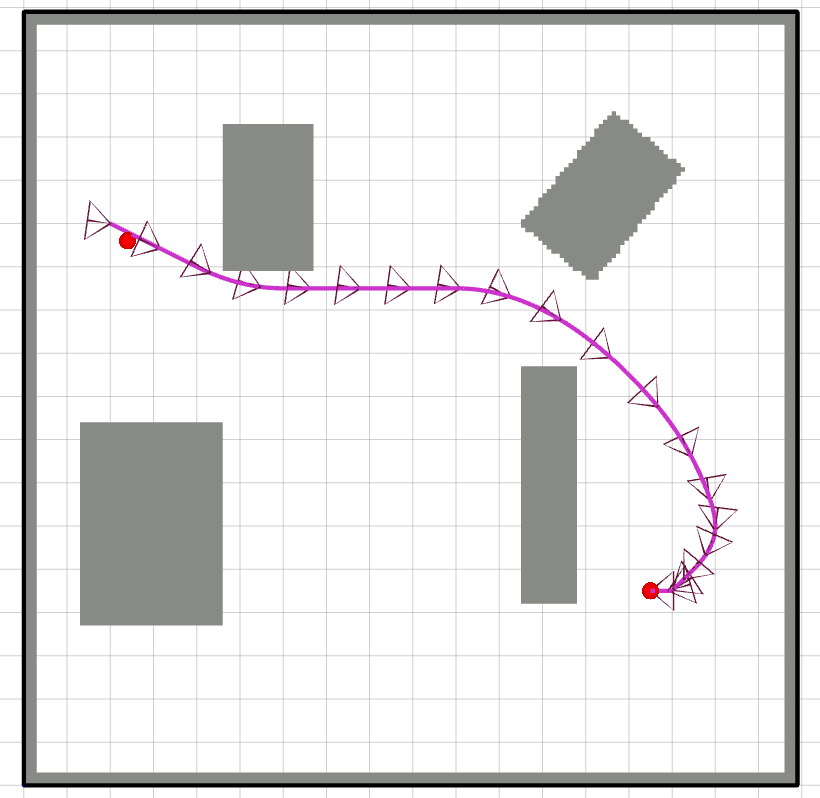}};
\node at (100pt,25pt) {start};
\node at (20pt,95pt) {goal};
\draw[->](100pt,33pt)--(113pt,33pt);
\end{tikzpicture}
        \caption{$\rho_\psi = 1,~ \theta = \pi/2$.}\label{fig:yaw3}
    \end{subfigure}
    \caption{Planning from a start that faces towards right to a goal with a non-zero initial velocity (black arrow), with yaw constraint. We draw the desired yaw as a small triangle at the corresponding position. As we adjust the parameters $\rho_\psi$ and $\theta$, the desired yaw along the planned trajectory follows different profiles.\label{fig:yaw}}
\end{figure}
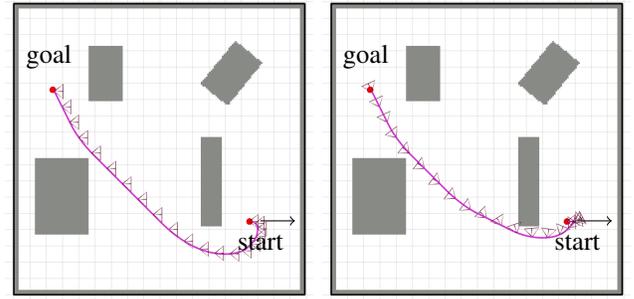
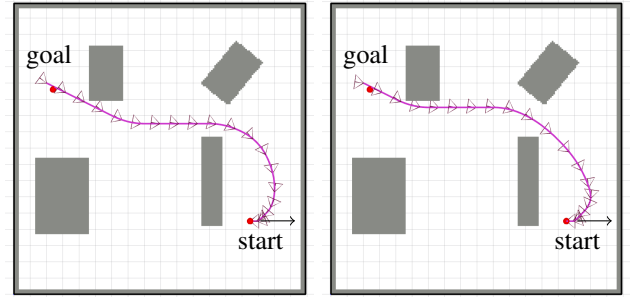

\autoref{fig:yaw} shows the planning results from solving \prob\ref{prob:3} with different parameters $\rho_\psi,~\theta$: in (a), we ignore the FOV constraint; in (b), we ignore the hard FOV constraint by setting $\theta = 2\pi$; in (c), we ignore the soft FOV constraint on $J_\psi$; and in (d), we consider both soft and hard constraints. Obviously, trajectories in (a) and (b) are not safe to follow since the robot is not always moving in the direction that the obstacle is visible within the sensor's FOV. The trajectory in \autoref{fig:yaw3} is desirable as its yaw is always following the velocity direction. Besides, even though the shapes of all the trajectories in \autoref{fig:yaw} look the same, the trajectories in (c) and (d) have longer duration since the robot needs to rotate to align the yaw along the trajectory at the beginning.

\subsection{Experimental Results}
\begin{figure}[tb]
  \centering
   \begin{subfigure}[b]{0.32\linewidth}
        \includegraphics[width=\linewidth]{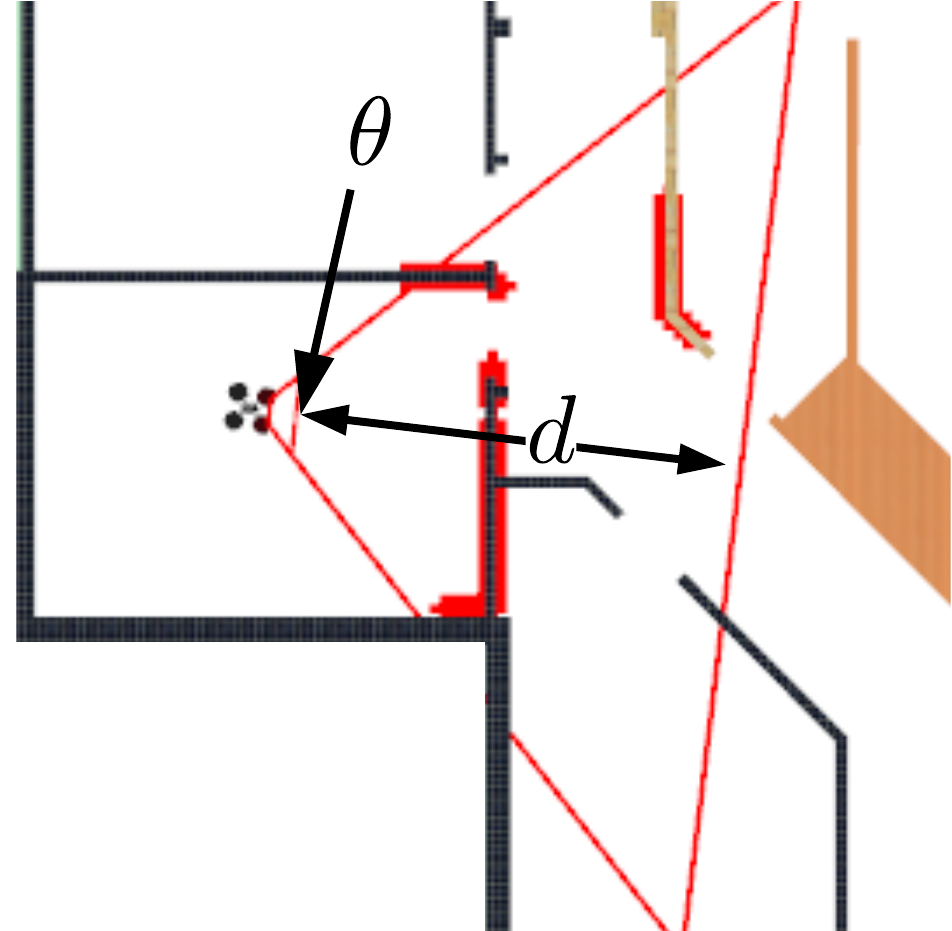}
        \caption{Sensor model.}\label{fig:sensor}
    \end{subfigure}
       \begin{subfigure}[b]{0.32\linewidth}
        \includegraphics[width=\linewidth]{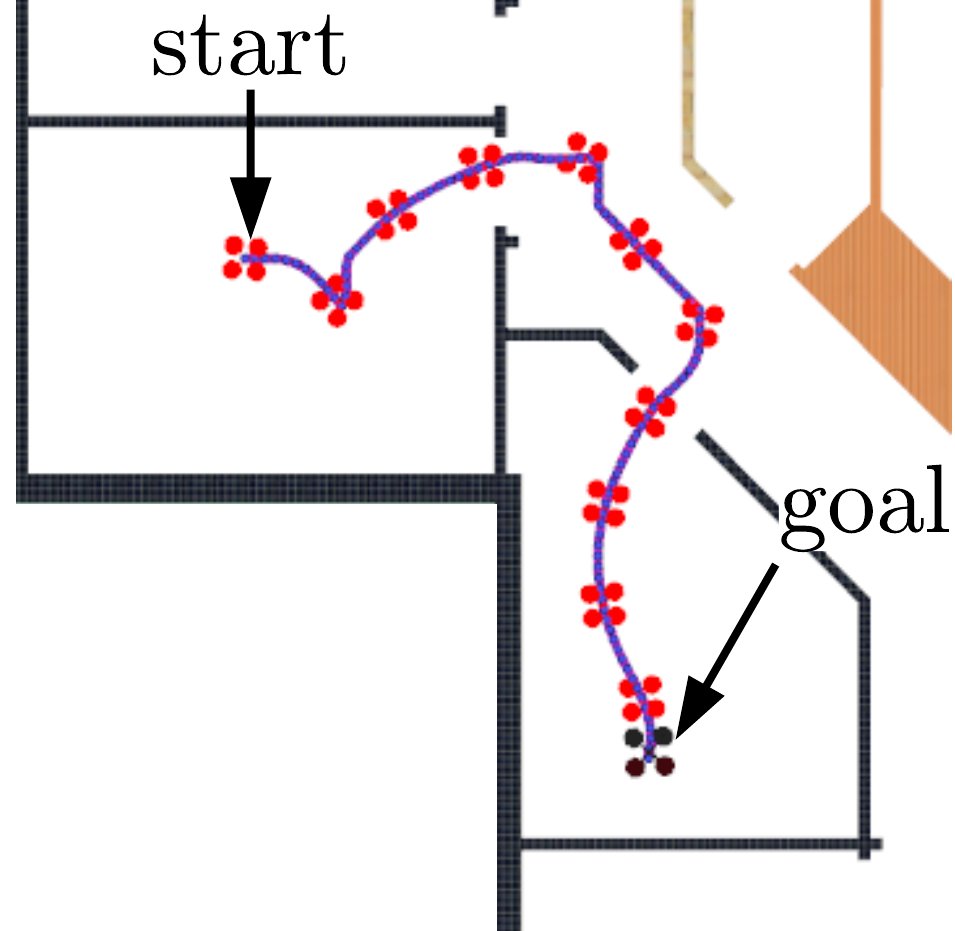}
         \caption{2D Navigation.}\label{fig:yaw_exp}
    \end{subfigure}
           \begin{subfigure}[b]{0.32\linewidth}
        \includegraphics[width=\linewidth]{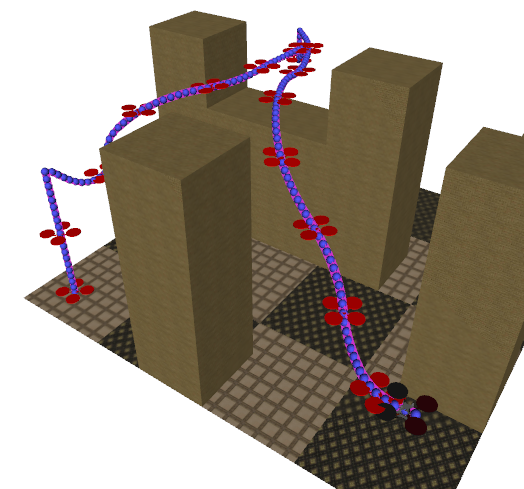}
         \caption{3D Navigation. }\label{fig:yaw_exp2}
    \end{subfigure}
   \caption{Navigation of a quadrotor equipped with an RGB-D camera in an office environment. The red triangle in the left image indicates the sensor's FOV $\theta$ and sensing range $d$. The red cells in (a) stand for the points detected by the sensor. The trajectory in (b) and (c) shows the quadrotor approaching the goal with changing yaw using the proposed method. \label{fig:traj_yaw_exp}}
\end{figure}

The yaw constraint can be used with the APF constraint described in \autoref{sec:disturbance} by adding $J_c$ in the cost function of~\prob\ref{prob:3} as
\begin{equation}
J_q + \rho_T T + \rho_\psi J_{\psi} +\rho_c J_c.
\end{equation}
The solution to this modified problem satisfies the requirements of directional movement and safety. Similar to \autoref{sec:exp_perturb}, we use this planner to generate and re-plan trajectories from start to goal in both 2D and 3D environments (\autoref{fig:traj_yaw_exp}). The environment is initially unknown, and the robot uses its onboard depth sensor with FOV $\theta$ to detect obstacles. To be able to plan trajectories reaching the goal, we need to treat unexplored space as free space. This greedy assumption introduces the risk that the trajectory could potentially crash the robot into hidden obstacles that are outside of the sensor's FOV. Our planner is able to generate yaw movements along the trajectory such that the robot is always moving into the region in the sensor's FOV. Therefore, the robot is able to avoid hitting hidden obstacles and reach the goal safely.

\section{Planning in Dynamic Environments}
\label{sec:moving_obstacles}
All planning problems addressed so far involved a static map. Ensuring completeness in environments with mobile obstacles is much harder~\cite{reif1994motion}. Existing planning methods based on fast re-planning including~\cite{hsu2002randomized, koenig2002d} or safe interval~\cite{sipp} are neither complete nor efficient. Reactive collision avoidance using the \emph{velocity obstacle} (VO)~\cite{fiorini1998motion, van2011reciprocal} discards the global optimality and completeness to gain the guarantee of flight safety and real-time computation. However, these VO based frameworks assume a simple straight line path with constant velocity and cannot be used to follow a dynamically feasible trajectory.

In this section, we directly solve the planning problem in a dynamic environment using our search-based framework which is resolution optimal and complete. To ensure flight safety, the robot needs to frequently re-plan since the information of surrounding moving obstacles is constantly updated. We model a moving obstacle as the \emph{linear velocity polyhedron} (LVP) in $\mathbb{R}^m$ whose position and velocity are observable. In fact, a linear model for a moving obstacle is only an approximation of its motion in the general case. To increase the success of future re-plans, we inflate LVP with respect to time. In the meanwhile, to avoid wasting time searching over the same region repeatedly, we use an incremental trajectory planning approach based on Lifelong Planning A* (LPA*)~\cite{koenig2004lifelong}. The proposed planner can further be developed for planning for multi-robot systems, in which the inter-robot collision avoidance is guaranteed.

\subsection{Model of Moving Obstacles}
\label{sec:model_obstacles}
Consider a single moving obstacle and suppose that it is represented as LVP which is a convex polyhedron $c$ in $\mathbb{R}^m$ with velocity $v_c$ (no rotation). We first show that the collision between a polynomial trajectory $\Phi$ and $c$ can be checked by solving for roots of a polynomial. Then, we describe the model of motion uncertainty of the LVP in re-planning.

\subsubsection{Collision Checking}
Denote a half-space in $\mathbb{R}^m$ as $h = \{p~|~a^\T p \leq b,~p\in \mathbb{R}^m \}$. The intersection of $M$ half-spaces gives a convex polyhedron, $c = \bigcap_{j=0}^{M-1} h_j = \{p~|~\mathbf{A}^\T p \leq \mathbf{b},~p \in\mathbb{R}^m\}$, where $a_j$ corresponding to $h_j$ is the $j$-th column of matrix $\mathbf{A}$ and $b_j$ is the $j$-th element of vector $\mathbf{b}$.  If a polynomial trajectory $\Phi$ defined by~\eqref{eq:traj} collides with $c$, we must have one of its trajectory segments $\Phi_i$ intersect $c$ in the time interval $[t_i, t_{i+1}]$. It can be verified by finding roots of the polynomial function of time $a_j^\T \Phi_i(t) = b_j$: if there exists a root $t_c$ located in the interval $[0, t_{i+1}-t_i]$ and the intersecting point $\Phi_i(t_c)$ (\ie $\Phi(t_i+t_c)$) is on the boundary of $c$, we claim that $\Phi$ collides with $c$.

\begin{proposition}\label{prop:collision}
A trajectory segment $\Phi_i(t)$ intersects a polyhedron $c$ composed of half-spaces described by $\mathbf{A}$ and $\mathbf{b}$ if and only if
\begin{equation}\label{eq:collision}
  \exists~t_c \in [0, \Delta t_i] \quad \text{s.t.} \quad \mathbf{A}^\T \Phi_i(t_c) \leq \mathbf{b}
\end{equation}
\end{proposition}



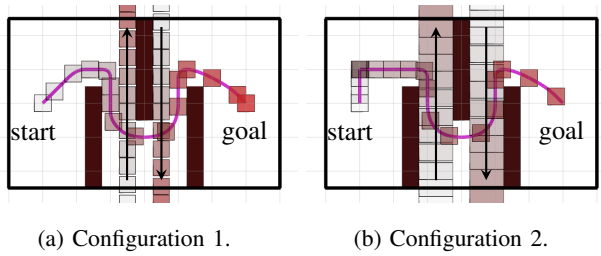
\begin{figure}[htp]
	\centering
	\begin{subfigure}[b]{0.23\textwidth}
\begin{tikzpicture}
\node[above right] (img) at (0,0) {\includegraphics[width=1.00\textwidth]{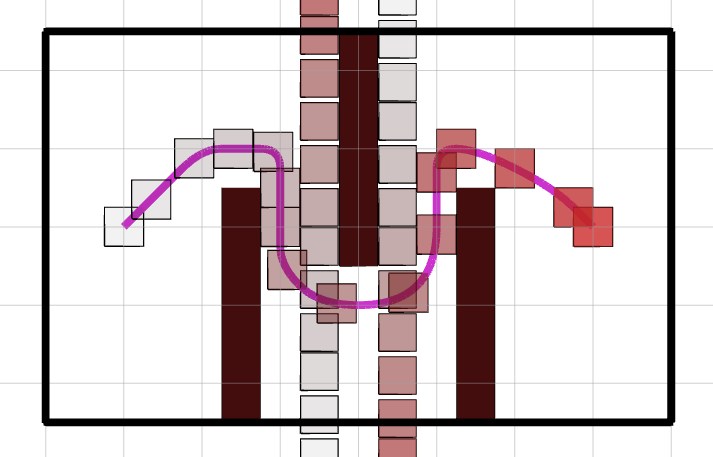}};
\node at (20pt,30pt) {start};
\node at (100pt,30pt) {goal};
\draw[->, >=stealth, thick]  (55.5pt, 12pt) --  (55.5pt, 70pt);
\draw[->, >=stealth, thick]  (68.5pt,70pt) -- (68.5pt, 12pt);
\end{tikzpicture}
        \caption{Configuration 1.}
     \end{subfigure}
     \begin{subfigure}[b]{0.23\textwidth}
     \begin{tikzpicture}
\node[above right] (img) at (0,0) {\includegraphics[width=1.00\textwidth]{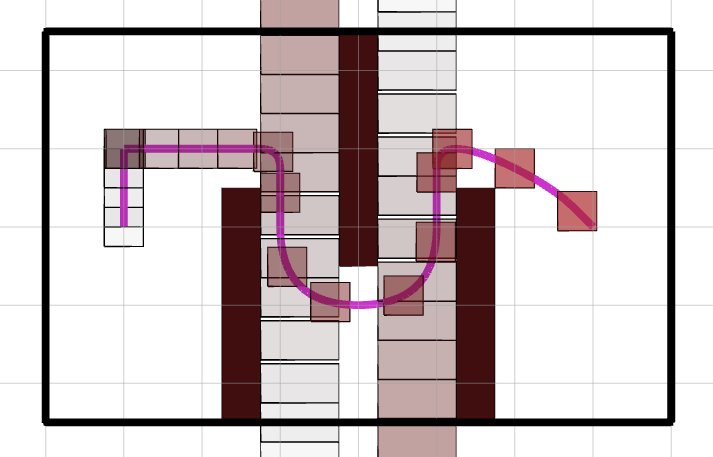}};
\node at (20pt,30pt) {start};
\node at (100pt,30pt) {goal};
\draw[->, >=stealth, thick]  (52.5pt, 12pt) -- (52.5pt, 70pt);
\draw[->, >=stealth, thick] (71.5pt,70pt) -- (71.5pt, 12pt);
\end{tikzpicture}
        \caption{Configuration 2.}
     \end{subfigure}
     \caption{Planning with linearly moving obstacles. We use different transparencies to represent positions of moving obstacles and robot at different time stamps. The moving obstacles in configuration 2 is wider than configuration 1, thus the planned trajectory in configuration 2 let the robot wait for the first obstacle passing through the tunnel instead of entering the tunnel in parallel. \label{fig:moving_obstacles}}
\end{figure}

Note that $a_j^\T$ is the outward normal of the half-space $h_j$ and it is invariant with respect to time since we assume that the obstacles do not rotate. $b_j$ is time-varying if $v_c$ is non-zero. Denote $h_j^0 = \{p~|~a_{j, 0}^{\T} p \leq b_{j,0}\}$ as the initial half-space $h_j^0 = h_j(t = 0)$, we have
\begin{equation}\label{eq:obs}
a_j(t) = a_{j, 0},~b_j(t) = b_{j,0} + a_{j,0}^\T\cdot v_ct.
\end{equation}
Since we assume that $v_c$ is constant for each planning interval, $b_j(t)$ is a time-parameterized polynomial function. Therefore, we are still available to solve roots from the polynomial function in~\eqref{eq:collision}.~\autoref{fig:moving_obstacles} shows the planning results in two configurations with LVPs. For better visualization, the animation of robot following planned trajectories in corresponding dynamic environments is shown in the accompanying video.

\subsubsection{Uncertainty of Linear Polyhedron}
Since the movement of a moving obstacle is unpredictable, our LVP model is an optimistic prediction for the purposes of re-planning. To address this problem, we use a simple strategy similar to~\cite{van2006planning} that grows the obstacle's geometry: shift all the half-spaces in the direction of the outward normal (namely, $a$) with certain speed $v_e > 0$. As a result,~\eqref{eq:obs} is modified as:
\begin{equation}\label{eq:obs2}
a_j(t) = a_{j, 0},~b_j(t) = b_{j,0} + (a_{j,0}^\T\cdot v_c + \|a_{j,0}\| v_e)t.
\end{equation}

\begin{figure}[tb]
	\centering
	\begin{subfigure}[b]{0.23\textwidth}
        \includegraphics[width=\textwidth]{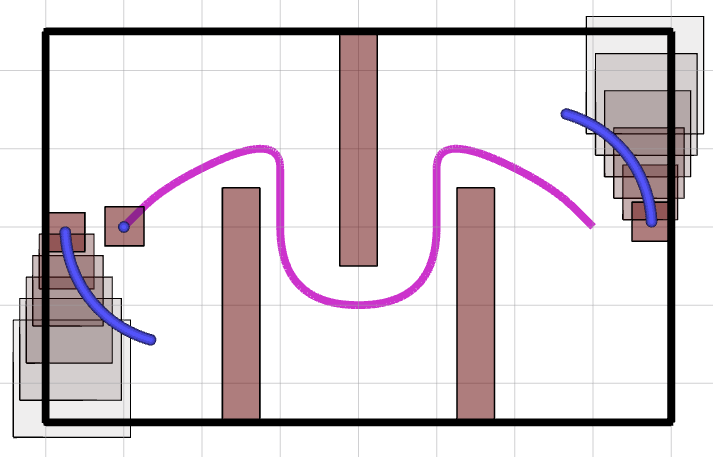}
        \caption{Plan epoch 0.}
     \end{subfigure}
     \begin{subfigure}[b]{0.23\textwidth}
        \includegraphics[width=\textwidth]{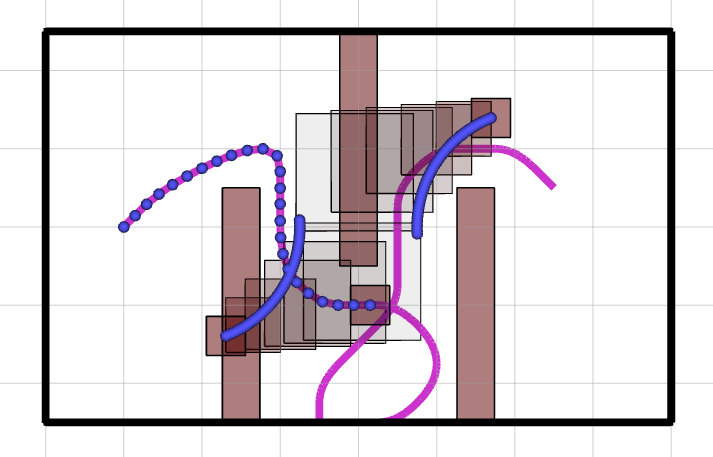}
        \caption{Plan epoch 12.}
     \end{subfigure}
     \begin{subfigure}[b]{0.23\textwidth}
        \includegraphics[width=\textwidth]{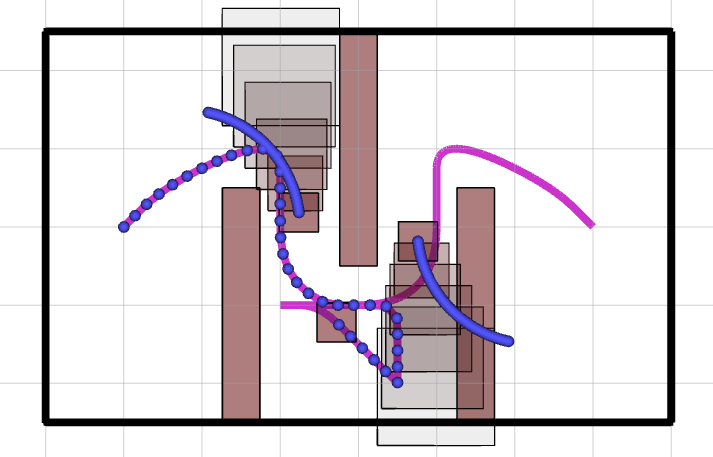}
        \caption{Plan epoch 15.}
     \end{subfigure}
     \begin{subfigure}[b]{0.23\textwidth}
        \includegraphics[width=\textwidth]{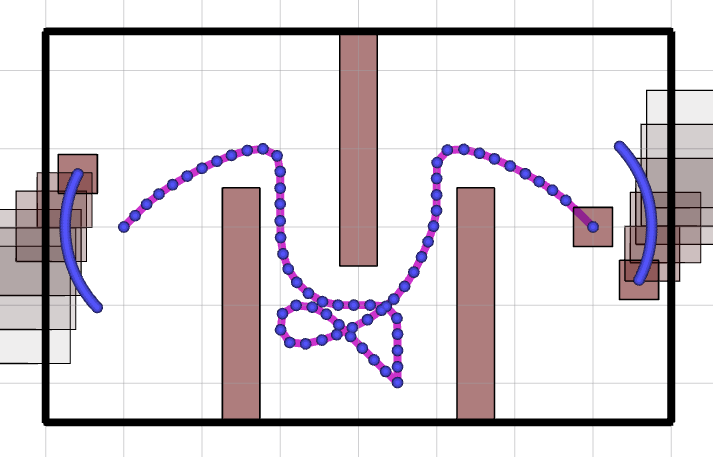}
        \caption{Plan epoch 20.}
     \end{subfigure}
	\caption{Re-planning with moving obstacles with uncertainty. The blue splines show the future trajectories of moving obstacles which are unobservable. The stacked transparent rectangles indicate the evolution of the moving obstacles as predicted by~\eqref{eq:obs2}, which are used in re-planning. \label{fig:moving_obstacles2}}
\end{figure}

Substituting~\eqref{eq:obs2} into~\eqref{eq:collision}, we can still get a polynomial function to check for collision. An example of growing obstacles is illustrated in~\autoref{fig:moving_obstacles2} where the robot constantly re-plans at \SI{1}{\hertz}. The robot is able to avoid the non-linearly moving obstacles with the proposed linear model in~\eqref{eq:obs2} with a properly selected $v_e$.

\subsection{Problem Formulation}
For a general planning problem in the environment that has both static and moving obstacles, we separate the collision checking in two workspaces: a static workspace $\mathcal{X}_s$ and a dynamic workspace $\mathcal{X}_d(t)$. $\mathcal{X}_s$ can be represented by a standard map which contains a collision-free subset $\mathcal{X}^{free}_s$ and an occupied subset $\mathcal{X}^{obs}_s = \mathcal{X}_s \setminus \mathcal{X}^{free}_s$. $\mathcal{X}_d(t)$ is a time-varying set whose occupied subset consists of $K$ moving obstacles $\mathcal{X}_d^{obs}(t) = \bigcup_{k=0}^{K-1} c_k(t)$. Denote the free subset as $\mathcal{X}^{free}_d(t) = \mathcal{X}_d(t) \setminus \mathcal{X}^{obs}_d(t)$, the original collision constraint $x(t) \in \mathcal{X}^{free}$ in~\prob\ref{prob:1} is re-written as:
\begin{equation}
x(t) \in \mathcal{X}_s^{free}, ~\text{and}~ x(t) \in \mathcal{X}^{free}_d(t).
\end{equation}
Since the collision checking is a function of time, the lattice state in~\prob\ref{prob:1} should be augmented by the corresponding time stamp. In such case, a maximum planning horizon $T_{max}$ is the criterion to determine if the search should be terminated or not. Otherwise, if the goal is occupied permanently by moving obstacles, the planner will keep expanding the same state at different time stamps.

\subsection{Incremental Trajectory Planning}\label{sec:itp}
A planned trajectory needs to be updated when new information of moving obstacles is updated in order to guarantee safety and optimality. Re-planning from scratch every time is not efficient since we may waste time searching places that are already explored in previous planning epochs. To leverage incremental search techniques for dynamic systems, we replace the A*  with Lifelong Planning A* (LPA*)~\cite{koenig2004lifelong}. For searching with motion primitives, an additional \emph{graph pruning} process is necessary to maintain the correctness and optimality of the planning results. By combining LPA* and \emph{graph pruning}, we can efficiently solve the re-planning problem in a dynamic environment.

\subsubsection{LPA* with Motion Primitives}
It is computationally expensive to construct the graph in free space due to feasibility checking of edges. Thus, if we are able to reuse the graph from the previous plan and only update costs of affected edges as the map changes, we are able to save a significant amount of time in the new planning query. D*~\cite{stentz1994optimal}, D* Lite~\cite{koenig2002d} are popular incremental graph search algorithms that have been widely used in real-time re-planning. However, these algorithms are not suitable for state lattice search due to the fact that the induced lattice almost never hits the exact goal state. Thus the trajectory planned from the goal state to start (as done in D* and D* Lite) will not be able to exactly reach the start state. Besides, in the navigation task in unknown environments, instead of spending efforts on exploring regions that are far away from the robot's current location, we are more interested in searching its nearby region. Therefore, LPA* is a better choice for fast re-planning that plans the same optimal trajectory as A* but expands much fewer states.

In LPA*, once the graph has local inconsistency due to a map update, the algorithm will re-expand the affected states until all the locally inconsistent states become consistent. To update edge costs, we need to check the related edges in the graph according to the latest map. If the map is incrementally updated by a relative small amount, this update is sufficiently fast. More details about LPA* can be found in the original paper~\cite{koenig2004lifelong}.


\subsubsection{Graph Pruning}
When the robot starts following the planned trajectory, the start state for the new planning moves to the successor of the previous start state along the planned trajectory. Now, since the start state for the graph has been changed, the start-to-state cost of all the graph vertices need to be updated. Since our graph is directional (edges are irreversible), a large portion of the existing graph that originated from the old start state becomes unreachable. Hence, it is important to prune an existing graph and update the start-to-state cost of the remaining states according to the new start state. In this step, we do not need to explore new states, check for collision against map or calculate any heuristic values, thus it is very fast.

\subsubsection{Run Time Analysis}
\begin{figure}[tb]
\centering
       \begin{subfigure}[b]{0.49\textwidth}
        \includegraphics[width=\textwidth]{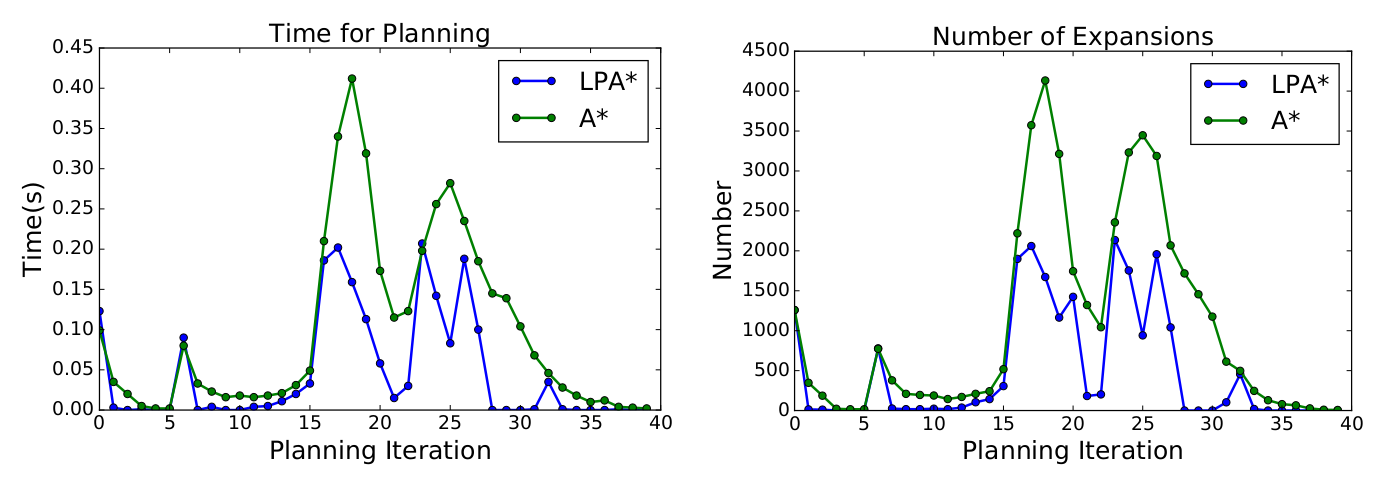}
    \end{subfigure}
   \caption{Comparison of efficiency between LPA* and A* in a navigation task of~\autoref{fig:moving_obstacles2}. \label{fig:compre}}
\end{figure}

We show the planning time and number of expansions of planners using A* and LPA* with \emph{graph pruning} in~\autoref{fig:compre}, from which we can clearly see that LPA* is more efficient than A* since it takes less planning time and requires fewer number of expansions over the whole mission.

\subsection{Multi-robot Planning}
\label{sec:multi_robot}
We consider a scenario where a team of homogeneous robots operates in a environment at the same time. We assume that some mission control algorithm such as exploration assigns a target to each of the robots. Thus, it is a decoupled problem in which individual robot plans its own trajectory. Different from existing works such as~\cite{van2009centralized, luna2011efficient, turpinIJRR2014}, we mainly focus on finding the optimal trajectory for robot without colliding with other robots. We show that the proposed framework can be used to plan trajectory for each robot by treating other robots as moving obstacles. Thus, we are able to perform either sequential or decentralized planning for multiple robots in the same workspace.

\subsubsection{Collision Checking between Robots}
In \autoref{sec:model_obstacles}, we modeled the obstacle as a linearly moving polyhedron in $\mathbb{R}^m$, which can be generalized for non-linear moving obstacles that follow piecewise polynomial trajectories. Denote $c_i(t)$ as the $i$-th robot configuration which is a \emph{non-linearly moving polyhedron} (NMP) in $\mathbb{R}^m$, it is represented as the robot geometry $c_{i,0} = \{p~|~\mathbf{A}_i^\T p \leq \mathbf{b}_i\}$ that centered at robot's center of mass following a trajectory ${}^i\Phi(t)$. Thus, it can be represented as:

\begin{equation}\label{eq:robot_model}
c_i(t) = \{p~|~\mathbf{A}_i^\T(p - {}^i\Phi(t)) \leq \mathbf{b}_i\}.
\end{equation}
For robot $i$ and $j$, they are not colliding with each other if and only if
\begin{equation}\label{eq:collision_checking}
{}^i\Phi(t) \cap [c_{i,0} \oplus c_j(t)] = \emptyset,
\end{equation}
where ``$\oplus$'' denotes the Minkowski addition. Constraint~\eqref{eq:collision_checking} can be verified by solving for roots of a polynomial equation similar to~\eqref{eq:collision}. For a team of robots, we can verify whether the $i$-th robot's trajectory is collision-free by checking~\eqref{eq:collision_checking} against all the other robots.

\subsubsection{Sequential Planning}
For a team of $Z$ robots, we can sequentially plan trajectory for robots from $0$ to $Z-1$ by assigning priorities to the robots. When planning for $i$-th robot, we only consider collision checking with robots that have higher priority than $i$. Equivalently, we need to verify the following equation for the $i$-th robot:
\begin{equation}
{}^i\Phi(t) \cap \bigcup_{j = 0}^{i-1} [c_{i,0} \oplus c_j(t)] = \emptyset.
\end{equation}

\begin{figure}[tb]
\centering
    \begin{subfigure}[b]{0.225\textwidth}
        \includegraphics[width=\textwidth]{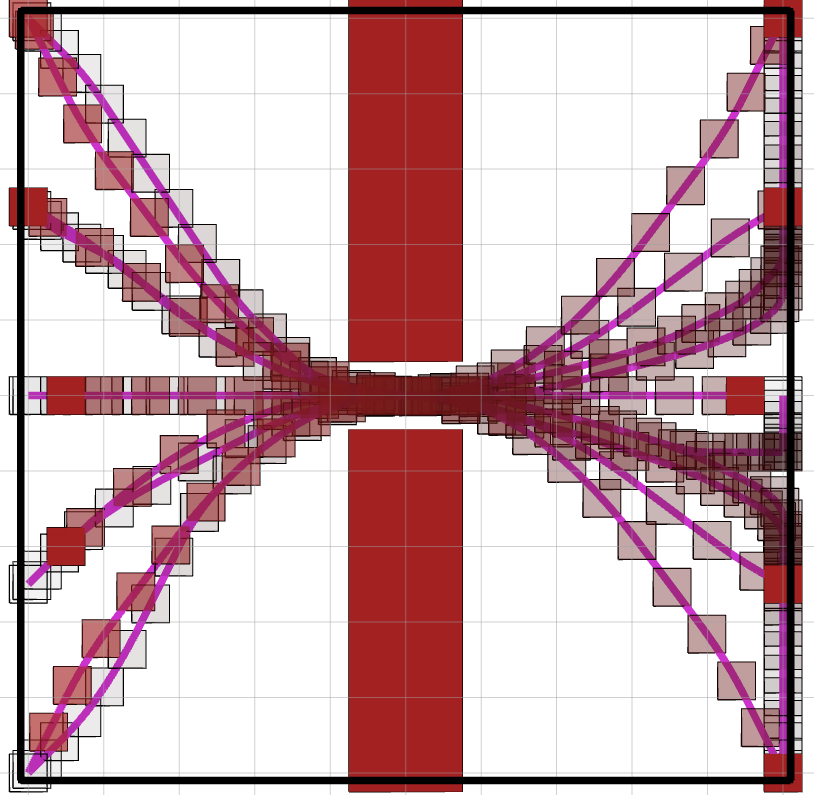}
        \caption{Tunnel configuration.}\label{fig:task1}
    \end{subfigure}
     \begin{subfigure}[b]{0.225\textwidth}
        \includegraphics[width=\textwidth]{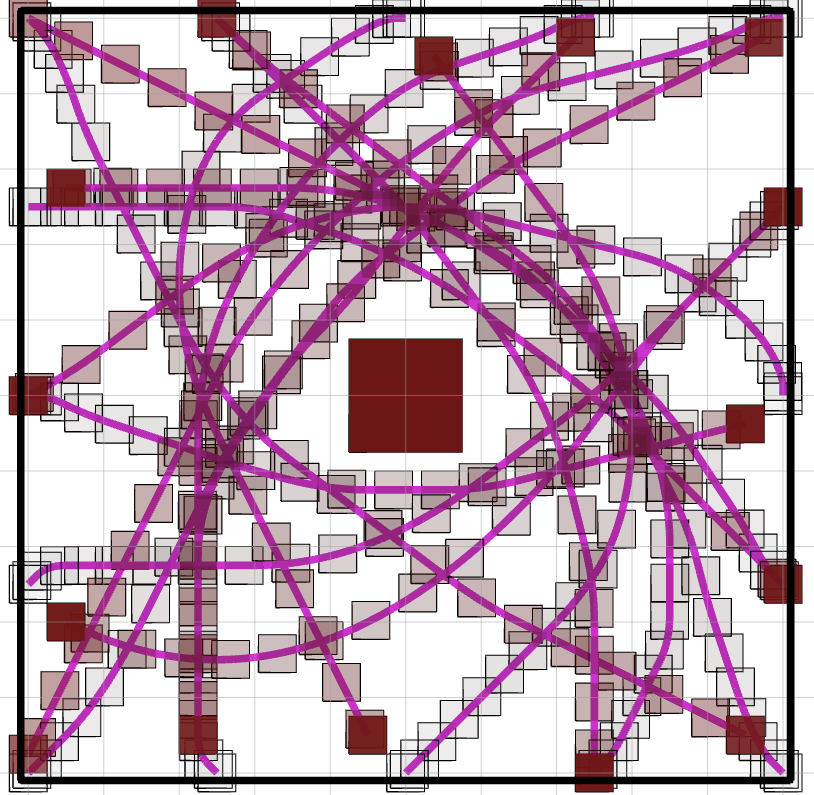}
        \caption{Star configuration.}\label{fig:task2}
    \end{subfigure}
   \caption{Example planning tasks for a multi-robot system. Blue rectangles represent the obstacles and robots' geometries. Magenta trajectories are the planning results from the sequential planning. \label{fig:tasks}}
\end{figure}

Sequential planning is able to guarantee inter-robot collision-free and find the optimal trajectory for each robot with respect to the priority.
The planning results for two navigation tasks are shown in~\autoref{fig:tasks}. Its computational complexity is polynomial, thus we are able to quickly plan the trajectories for the whole team.

\subsubsection{Decentralized Planning}
In the decentralized case, each robot re-plans at their own clock rate and there is no priority. It is practical to assume that the robot is able to share information about its current trajectory with other robots. For accurate collision checking, we also assume there is a global time frame and a local time frame for each robot representing its trajectory start time. Use $\tau$ and $t$ to represent the time in global time frame and local time frame respectively. The conversion between this two frames is simply $t = \tau - \tau^s$ where $\tau^s$ is the start time in global time frame of the trajectory. Thus, we formulate the collision checking for $Z$ robots in the decentralized manner as:
\begin{equation}
{}^i\Phi(\tau-\tau_i^s) \cap \bigcup_{j = 0}^{Z-1} [c_{i,0} \oplus c_j(\tau-\tau_j^s)] = \emptyset,
\end{equation}
or in local time frame as:
\begin{equation}\label{eq:collision_checking2}
{}^i\Phi(t) \cap \bigcup_{j = 0}^{Z-1} [c_{i,0} \oplus c_j(t+\tau_i^s-\tau_j^s)] = \emptyset.
\end{equation}
Here $\tau_i^s-\tau_j^s$ is a constant for the given trajectory pair ${}^i\Phi$ and ${}^j\Phi$. Since we plan for robot $i$ with the presence of robot $j$, $\tau^s_i - \tau^s_j \geq 0$ should always be true.

\begin{figure}[tb]
\centering
    \begin{subfigure}[b]{0.225\textwidth}
        \includegraphics[width=\textwidth]{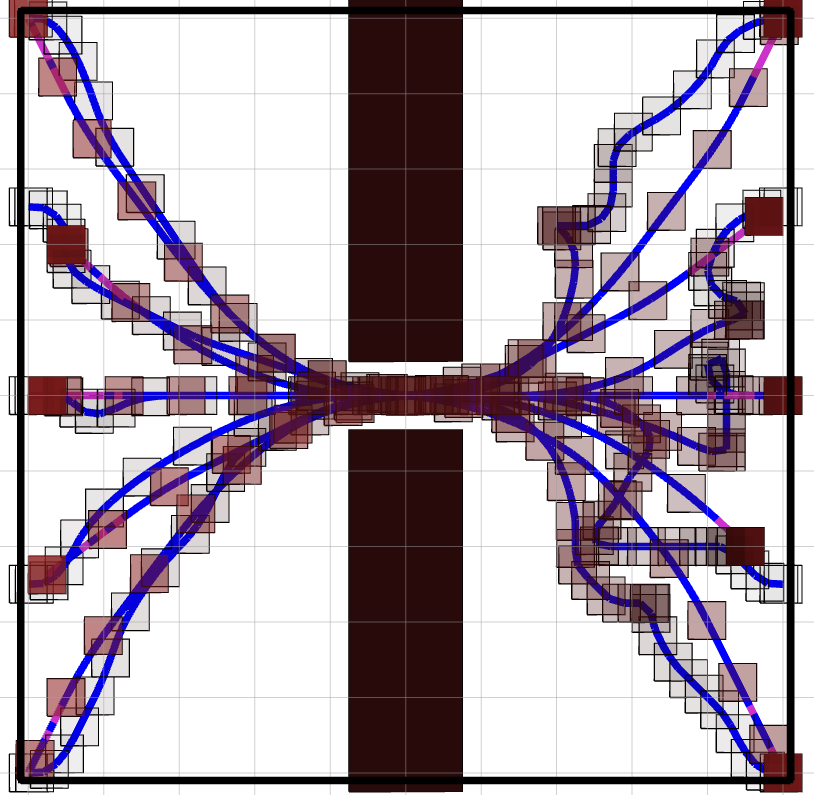}
        \caption{Tunnel configuration.}
    \end{subfigure}
     \begin{subfigure}[b]{0.225\textwidth}
        \includegraphics[width=\textwidth]{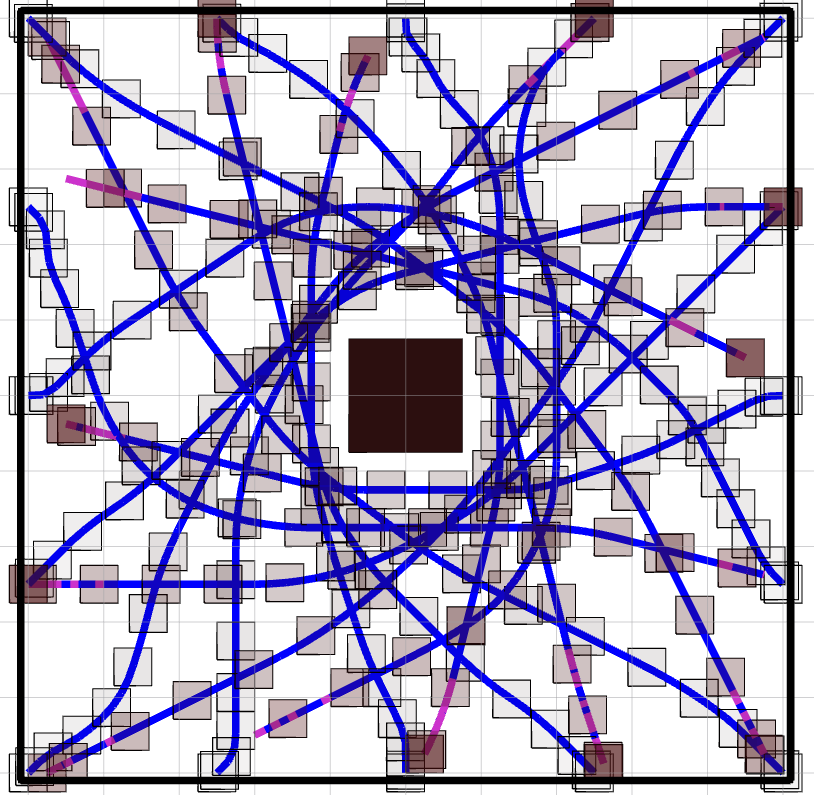}
        \caption{Star configuration.}
    \end{subfigure}
   \caption{Example planning tasks for a multi-robot system. Blue rectangles represent the obstacles and robots' geometries. Blue trajectories are the traversed trajectories of the team guided by the proposed decentralized planning. \label{fig:tasks2}}
\end{figure}

Denote ${}^jT$ as the duration of trajectory ${}^j\Phi$. Ideally, we use the whole trajectory ${}^j\Phi(t)$ from $t = 0$ to ${}^jT$ for collision checking when plan for robot $i$. However, we know that robot $j$ is also constantly re-planning, such that the future trajectory of $j$ can be meaningless to be considered in collision checking. We are able to improve the efficiency of collision checking in~\eqref{eq:collision_checking2} by setting a cutoff time $T_c$. Namely, we ignore the part of trajectory ${}^j\Phi(t)$ of other robot $j$ for the domain $t > {}^jT_c$. Consequently, as $T_c$ gets smaller, the computational time for inter-robot collision checking is also smaller. The smallest $T_c$ for a complete solution is determined by the system's dynamic constraints. For example, for a second order system that is constrained by maximum velocity $\bar{v}_{max}$ and acceleration $\bar{a}_{max}$, the smallest value of $T_c$ is the minimum time it takes to stop the robot from the maximum velocity. Thus, we set the value of $T_c$ for trajectory $\Phi_j$ as
\begin{equation}
{}^jT_c = \min\{\bar{v}_{max} / \bar{a}_{max}, {}^jT\}.
\end{equation}
To make the algorithm complete, we need to ignore robot $j$ for $t > {}^jT_c$ when planning for robot $i$ instead of treating it as a static obstacle.~\autoref{fig:tasks2} shows the results of two planning tasks using the decentralized planning with $T_c$.

\section{Conclusion}
\label{sec:conclusion}
In conclusion, we proposed a way to use a search-based motion planning method with motion primitives to solve planning problems with uncertainty in the model and with FOV constraints. The proposed planner is shown to be efficient and is resolution complete and optimal. We believe the proposed methodology has the potential to be used in a broad class of MAV navigation problems even in dynamically changing environments. We note that although the examples shown in this paper are mostly done in 2D environments, our framework has already been tested in 3D as well.

\balance
\bibliographystyle{bib/IEEEtran}
\bibliography{bib/references}

\end{document}